%% file: root.tex
\documentclass[letterpaper, 10 pt, conference]{ieeeconf}  

\usepackage[table,xcdraw]{xcolor}
\usepackage{multirow} 
\usepackage{amsthm}
\newtheorem{remark}{Remark}
\usepackage{amssymb}
\usepackage{amsmath}
\DeclareMathOperator*{\argmax}{arg\,max}
\usepackage{bm}
\usepackage[ruled,vlined,linesnumbered,noend]{algorithm2e}
\SetAlFnt{\small}\SetAlCapFnt{\small}\SetAlCapNameFnt{\small}
\SetInd{0.4em}{0.8em}

\SetCommentSty{BlueComment}
\SetKwComment{Comment}{\textcolor{blue}{$\triangleright$ }}{}\usepackage{amsmath}
\usepackage{graphicx}
\usepackage{caption}
\usepackage{subcaption}
\usepackage{hyperref}
\newenvironment{proof sketch}{\par\noindent\textit{Proof sketch.}\ }{\hfill$\square$\par}
\usepackage{booktabs} 
\usepackage{siunitx}  
\usepackage{makecell} 
\usepackage{graphicx}

\IEEEoverridecommandlockouts                              

\title{\LARGE \bf
Constant-Time Planning for Chaining Collision-free Motion to Manipulation Behaviors
}

\author{
  Nayesha Gandotra\thanks{$^{*}$ Equal contribution.}\thanks{$^{1}$ Robotics Institute, School of Computer Science, Carnegie Mellon University, Pittsburgh, PA, United States. \texttt{\{nayeshag, imishani, maxim\}@cs.cmu.edu}}$^{*,1}$,
  Itamar Mishani$^{*,1}$,
  Lai Yuan$^{1}$,
  Oren Salzman\thanks{$^{2}$ Department of Computer Science, Technion, Israel Institute of Technology, Haifa, Israel. \texttt{osalzman@cs.technion.ac.il}}$^{2}$,
  and Maxim Likhachev$^{1}$
}

\newtheoremstyle{mydefstyle}       
  {3pt}                            
  {3pt}                            
  {\itshape}                       
  {}                               
  {\bfseries}                      
  {.}                              
  {0.5em}                          
  {}                               

\theoremstyle{mydefstyle}
\newtheorem{mydef}{Definition}

\theoremstyle{plain}
\newtheorem{theorem}{Theorem}

\theoremstyle{mydefstyle}

\renewcommand{\baselinestretch}{0.933}

\begin{document}

\bstctlcite{IEEEexample:BSTcontrol}

\maketitle
\thispagestyle{empty}
\pagestyle{empty}

\input{sections/abstract}

\input{sections/introduction}

\input{sections/related_work}
\input{sections/problem_formulation}

\input{sections/algorithmic_approach2}

\input{sections/stat_guarantee}
\input{sections/experiments}
\input{sections/conclusion_future_work}

\bibliographystyle{IEEEtran}
\bibliography{root}

\end{document}

%% file: sections/abstract.tex
\begin{abstract}

Recent progress in contact-rich robotic manipulation has been striking, yet most deployed systems remain confined to simple, scripted routines.
One of the barriers is the lack of motion planning algorithms that can provide verifiable guarantees for safety, efficiency and reliability.
Constant-Time Motion Planning (CTMP) is a recent step toward such guarantees for collision-free motion in a priori known environments:: a preprocessing phase enables queries to be answered within a fixed, user-specified time budget (e.g., 10 milliseconds).
However, CTMP certifies only reachability---a binary predicate---and ignores the manipulation behavior that completes the task, which is increasingly stochastic (e.g., a learned skill) and whose success no single offline rollout can establish, let alone certify.
We introduce the Behavioral Constant-Time Motion Planner (B-CTMP), which extends CTMP to two-step manipulation tasks in semi-structured environments: a collision-free motion to a behavior initiation state, followed by execution of a behavior such as grasping or insertion.
B-CTMP departs from prior CTMP in two ways: neighborhoods are constructed in object-pose space rather than robot configuration space, and coverage is established by statistical certification rather than a reachability check.
A plan is cached only if repeated rollouts lower-bound its success rate above a user-specified threshold, and we prove these bounds hold simultaneously across the entire cache at a prescribed confidence level. For deterministic behaviors a single rollout suffices, recovering the binary check of prior CTMP as a special case.
We evaluate B-CTMP on three manipulation tasks---shelf picking, plug insertion, and wheel replacement---in simulation and on real robots. B-CTMP's certified plans succeed consistently where baselines fail during behavior execution, and it rejects infeasible object poses in constant time.

\end{abstract}

%% file: sections/introduction.tex
\section{Introduction}

Robotic arms have long automated tasks in highly structured domains such as automotive assembly lines and electronics manufacturing, where manipulators typically replay pre-recorded motions. This paradigm becomes fragile once variability is introduced: minor changes in the environment disrupt operation, and significant human effort is required for setup and maintenance. The fundamental limitation is that real-world manipulation requires collision-free motion planning and precise manipulation behaviors (e.g., grasping, insertion) to work together adaptively---when objects appear in different positions or orientations, the system must coordinate both, which fixed-motion approaches cannot do.

Despite the capabilities routinely demonstrated in research laboratories, this level of autonomy remains absent from industrial deployment. The gap is most pronounced in semi-structured environments such as warehouse shelf picking (e.g., Amazon fulfillment centers \cite{correll2016analysis}), bin picking in logistics, and precision assembly. A central challenge is the lack of verifiable guarantees on system performance---particularly safety, efficiency, and predictability---which practitioners in safety-critical and high-throughput applications require.

Constant-Time Motion Planning (CTMP) has been recently introduced as a promising framework for generating collision-free motion plans, in apriori known environments, within strict, user-defined time bounds \cite{ctmp, ConveyerCTMP, APP, mishani2024constant}. By leveraging offline computation, CTMP enables online planning of collision-free motions in constant time—often mere milliseconds (e.g., 10 milliseconds)—with guarantees of resolution-completeness within a predefined region of interest. 
However, existing CTMP methods do not explicitly address the manipulation aspect of the task—the part that involves interaction with the environment through manipulation behaviors such as grasping, insertion, or other contact-rich actions. This is a critical limitation, as the success of many real-world tasks depends on the precise execution of these behaviors at the goal.

\begin{figure}[t]
    \centering
    \includegraphics[width=0.94\columnwidth]{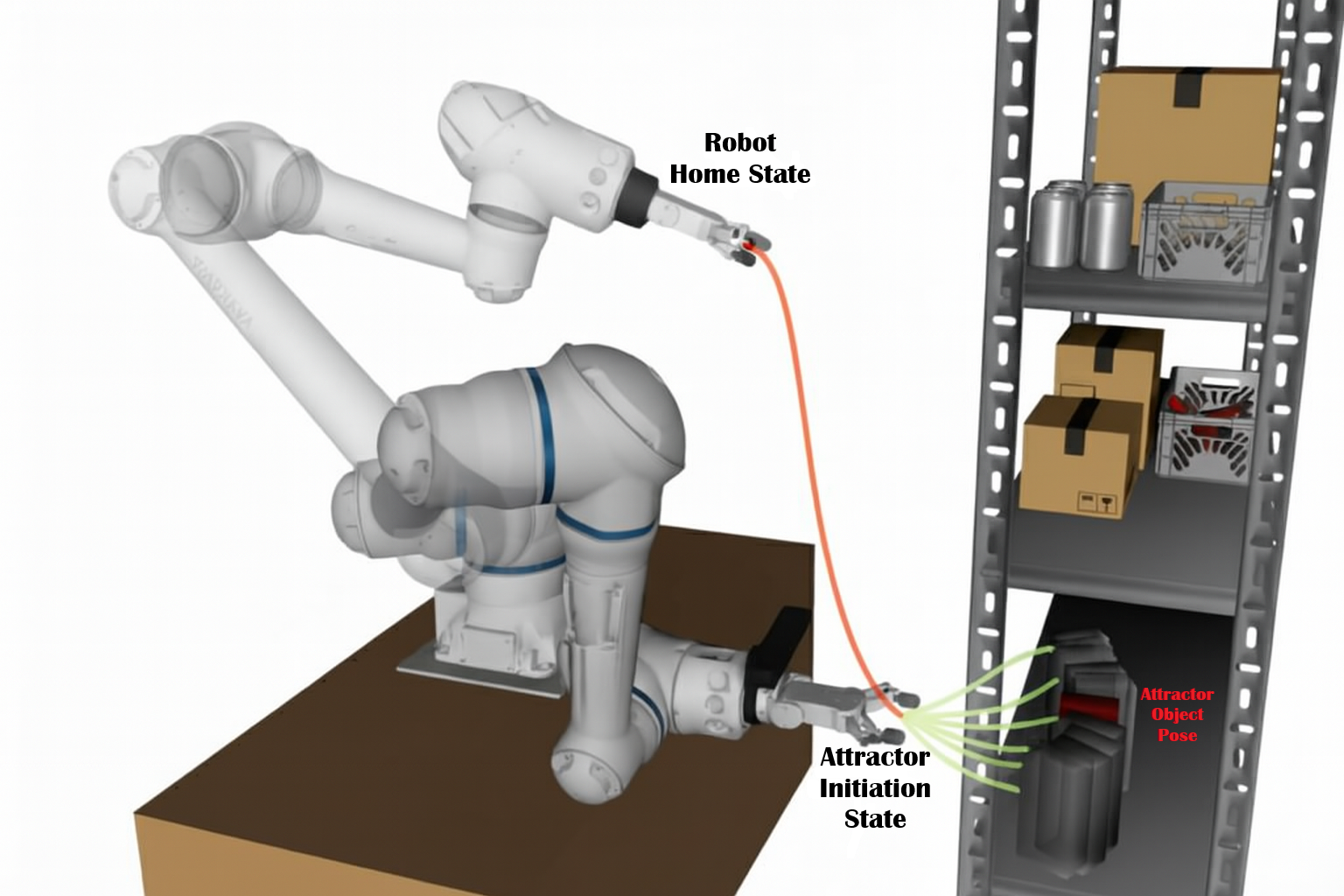}
    \caption{Shelf picking in industrial warehouse automation. Offline, B-CTMP caches a path from the robot's home state to an attractor initiation state (red), from which a behavior policy is executed to reach the target (green). Each cached plan is certified to meet a user-specified success rate, and is retrieved online in constant time.}
    \label{fig:cell}
\end{figure}

In this work, we bridge this gap by introducing the Behavioral Constant-Time Motion Planner (B-CTMP). B-CTMP incorporates manipulation behaviors directly into the preprocessing phase, enabling a two-step solution: (1) a collision-free motion to a behavior initiation state, and (2) execution of a manipulation behavior (e.g., grasping or insertion) to achieve the goal condition. Because such behaviors are increasingly \emph{stochastic}---as for skills learned via reinforcement or imitation learning---their success cannot be established by a single offline rollout, and B-CTMP therefore certifies it statistically. We evaluate B-CTMP on three manipulation tasks---shelf picking, plug insertion, and wheel replacement---in both simulation and on physical robots.

\textbf{Our main contributions are:}
\begin{itemize}
	\item We propose B-CTMP, a constant-time planning algorithm that validates manipulation behaviors during preprocessing, so its guarantee covers completing the task, not only reaching a configuration.
	\item We generalize coverage from deterministic reachability to the statistical certification of stochastic skills, and prove that B-CTMP is $(\alpha)$-complete: with confidence $1-\alpha$, every returned plan succeeds with a user specified success probability, with deterministic behaviors as a special case.
	\item We construct neighborhoods in object-pose space rather than robot configuration space, so a single cached plan covers a contiguous region of object poses.
\end{itemize}

%% file: sections/related_work.tex
\section{Related Work}

We organize the related work into into key areas that directly inform our approach: preprocessing-based methods for collision-free motion planning, and approaches for modeling manipulation behaviors and integrating them with planning.

\subsection{Motion Planning with Preprocessing}
Preprocessing (i.e., offline computations) has been a key component in the development of planning algorithms. The main purpose of preprocessing for collision-free motion planning is to efficiently compute data structures that enable fast real-time planning. 
The Probabilistic Roadmap (PRM) \cite{PRM} algorithm and its variants pioneered the approach of reducing the configuration space to a significantly smaller subset of states through roadmap construction, enabling fast online planning. 
However, PRM does not guarantee solution existence, as success depends on roadmap density and only provides asymptotic completeness guarantees \cite{PRM_guarantees}.

To address these limitations, recent approaches focus on constructing collision-free regions in the configuration space rather than sampling discrete states, making motion within these regions computationally efficient during online planning.
A prominent example of this region-based approach is Planning in Graphs of Convex Sets \cite{marcucci2023shortest, mp_aroundCS} which decomposes the configuration space offline into collision-free convex sets \cite{polyhedral-convex, np-convexsets}, enabling fast generation of smooth motion plans. 
These algorithms offer several advantages: they enable smooth trajectory generation through convex optimization and can handle complex geometric constraints naturally. 
Even though they do not guarantee time bounds on online planning duration, they are usually very fast. However, they do not take the post-motion manipulation behavior such as grasping into account.

An orthogonal line of work accelerates online planning through hardware parallelism rather than preprocessing. Vectorized and GPU-parallel planners such as VAMP \cite{thomason2024vamp} and cuRobo \cite{sundaralingam2023curobo} demonstrate collision-free planning in microseconds to milliseconds with no offline phase. Similar to GCS, these methods do not provide time bounds on online planning duration, but are usually very fast. Additionally, like other purely collision-free planners---they do not take the manipulation behavior itself into account. They are nonetheless complementary to ours: since B-CTMP treats the planner $\mathcal{P}$ as a black box, they can be used directly to accelerate preprocessing.

Constant-Time Motion Planning (CTMP)~\cite{ctmp, APP} addresses the time-bound limitation by providing provable guarantees for generating collision-free motion plans within user-defined time constraints in semi-static environments. The approach preprocesses a region-of-interest into sub-regions (neighborhoods), computing compact data structures that include representative paths for each region and potential functions that, when greedily followed (e.g., steepest descent), guarantee collision-free motion to the goal. Extensions include handling dynamic goal objects~\cite{ConveyerCTMP}, anytime planning~\cite{mishani2024constant}, continuous obstacle placements~\cite{cover}, and online adaptation~\cite{coad}. However, unlike the work in this paper, existing CTMP methods focus solely on collision-free motion and do not incorporate manipulation behaviors, making them less applicable  for contact-rich tasks.

\subsection{Manipulation Behaviors}
\label{rel:B}
A significant body of work has focused on designing \cite{mason2001mechanics, posa2013direct} and learning \cite{janner2022diffuser, zhou2024adaptive, aloha} manipulation behaviors. Planning with such behaviors typically involves defining key attributes for each, including initiation states (pre-conditions) and effects (post-conditions) \cite{sutton1999options, tamp, konidaris2014constructing, crosby2016skiros, kroemer2021review}, and integrating them with motion planning through a two-step \emph{independent} process: a collision-free motion reaches a state in the behavior's initiation set, after which the behavior is executed.

This separation places the burden on the motion planner to find, online, an initiation state from which execution will actually succeed. A robot may plan a collision-free path to a grasping pose whose approach trajectory is unsuitable for reliable grasp execution, given the object geometry or surrounding obstacles. No guarantee links the planned motion to successful behavior execution, and none of these approaches bound the time required to return such a plan.

%% file: sections/problem_formulation.tex
\section{Preliminaries}
\label{Prob}
\label{sec:preliminary}

We consider a robot $\mathcal{R}$ with state space $\mathcal{X}$ operating in a semi-structured environment. Let $\mathcal{W} \subseteq \mathit{SE}(3)$ denote the \emph{object state space}; i.e, the set of possible poses of the object to be manipulated. We equip $\mathcal{W}$ with a task-dependent distance metric $d : \mathcal{W} \times \mathcal{W} \to \mathbb{R}_{\ge 0}$.
A \emph{manipulation behavior} $\sigma$ is defined by the tuple
\[
\sigma = \big(\pi_\sigma,\; I_\sigma,\; p_\sigma,\; \textsc{GetInitStates}\big),
\]
where $\pi_\sigma$ is the behavior's policy, which may be \emph{deterministic} (e.g., Jacobian control) or \emph{stochastic} (e.g., learned via reinforcement or imitation learning); $I_\sigma(w) \subseteq \mathcal{X}$ is the \emph{initiation set} for object state $w \in \mathcal{W}$, i.e., the collision-free robot states within the domain of $\pi_\sigma$ from which it can be executed; $p_\sigma(s, w) \in [0,1]$ is the probability that rolling out $\pi_\sigma$ from $s \in I_\sigma(w)$ succeeds for object state $w$, with deterministic behaviors corresponding to the special case $p_\sigma \in \{0,1\}$; and $\textsc{GetInitStates}(w)$ returns a finite subset $\mathcal{S}(w) \subseteq I_\sigma(w)$ of candidate initiation states.

We let $\mathcal{G} \subseteq \mathcal{W}$ denote a region-of-interest (RoI), representing the region where the target object might be located. Each RoI may potentially consist of a set of disjoint regions that we call \textit{local-RoIs} $(\mathcal{G}_i)$---that is, $\mathcal{G} = \bigcup\limits_{i=1}^n \mathcal{G}_i$. Given object state $w_g \in \mathcal{G}$, our objective is to plan a full solution $\xi = (\tau, \pi_\sigma)$ which consists of the collision-free motion plan $\tau$ and the invocation of the behavior policy $\pi_\sigma$, such that the combined plan ensures both safe motion of the robot and, with quantified reliability, successful completion of the manipulation task. To make ``quantified reliability'' precise, we fix a user-specified acceptance threshold $\lambda \in [0,1]$ on the behavior success rate and define:


\begin{mydef}[$\lambda$-Feasibility]
\label{def:lambda_feasible}
An object state $w \in \mathcal{W}$ is \emph{$\lambda$-feasible} if there
exists a robot initiation state $s_i \in \mathcal{S}(w)$ that is reachable from
$s_{\text{home}}$ via a collision-free path and satisfies
$p_\sigma(s_i, w) \ge \lambda$.
\end{mydef}

\noindent We write $\mathcal{G}^\lambda \subseteq \mathcal{G}$ for the set of $\lambda$-feasible object states in the RoI. The remaining states admit no plan at threshold $\lambda$; for example, an object may be placed where every grasp pose collides with the shelf, or where reaching the target port drives the arm into a singularity.



For $\lambda$-feasible object states, we require a planning-time guarantee:

\begin{mydef}[Constant-Time Planner]
\label{def:constant_time}
A planner is \emph{constant-time} if, for every $w_g \in \mathcal{G}^\lambda$, it returns a solution $\xi = (\tau, \pi_\sigma)$ for
$w_g$ within a time bound $T_{\text{bound}}$.
\end{mydef}

Achieving this by precomputing a path for every initiation state of every
object pose would be prohibitively expensive in memory. Our approach instead
exploits the fact that a single initiation state can serve many object
states, which we formalize through \emph{coverage}.

\begin{mydef}[$\lambda$-Coverage]
\label{def:lam_cover}
An initiation state $s_i$ \emph{$\lambda$-covers} an object state $w$ if $p_\sigma(s_i, w) \ge \lambda$.
\end{mydef}

Coverage lets us adapt the notion of \emph{neighborhood} from prior CTMP work
\cite{mishani2024constant} to regions of interest defined in $\mathcal{W}$:

\begin{mydef}[Neighborhood]
\label{def:neighborhoods}
A \emph{neighborhood} $n(s_i) \subset \mathcal{W}$ is a set of object states,
each $\lambda$-covered by a common initiation state $s_i$.
\end{mydef}
In practice a neighborhood is stored compactly as a ball in $\mathcal{W}$: an \emph{attractor} object state $w_{attr}$ and radius $r$ such that every $w$ with $d(w, w_{attr}) \le r$ is $\lambda$-covered by $s_i$ (Fig.~\ref{fig:neighbourhoods}). The metric $d$ thus defines both the radii computed offline and the containment test $d(w_g, w_{attr}) \le r$ evaluated online; we specify it per task in Section~\ref{sec:experiments}.

Coverage establishes a many-to-many relationship between robot and object
states: each object state $w_g \in \mathcal{G}$ may admit multiple initiation
states in $I_\sigma(w_g)$, while each initiation state $s_i$ may
$\lambda$-cover multiple object states through its neighborhood $n(s_i)$. We
exploit this by selecting a compact set of initiation states
$\widetilde{\mathcal{S}} =\{s_1, \dots, s_K\}$ whose neighborhoods collectively
span the RoI,
$
\mathcal{G}^\lambda \subseteq \bigcup_{k=1}^{K} n(s_k).
$
This allows us to develop a \textit{constant time planner} for all \textit{$\lambda$-feasible} goal states with a memory footprint that scales with $K = |\widetilde{\mathcal{S}}|$ rather than with
$|\mathcal{G}|$. For stochastic behaviors, $p_\sigma$ is unknown and can only be estimated from
finitely many rollouts, so $\lambda$-coverage can never be established
exactly. In Section~\ref{sec:stat_guarantee}, we develop the statistical
machinery that allows us to certify coverage with high confidence.

%% file: sections/algorithmic_approach2.tex
\section{B-CTMP: Algorithmic Approach}
\label{alg_app}

In the following sections, we outline the two key phases of the algorithm--the offline preprocessing phase and the online query phase--and demonstrate how this approach provides memory efficiency and certified execution guarantees.

\subsection{Preprocessing Phase}
\label{alg_app:preprocess}

To guarantee plan retrieval within $T_{\text{bound}}$ online, we offload expensive planning and behavior simulation to an offline preprocessing phase. Given $s_{home}$, $\sigma$, and a collision-free motion planner $\mathcal{P}$, the goal of preprocessing is to compute a compact set $\widetilde{\mathcal{S}}$ of initiation states whose neighborhoods cover $\mathcal{G}^\lambda$, together with a collision-free path from $s_{home}$ to each.

\begin{figure}[t]
    \centering
    \includegraphics[width=\columnwidth]{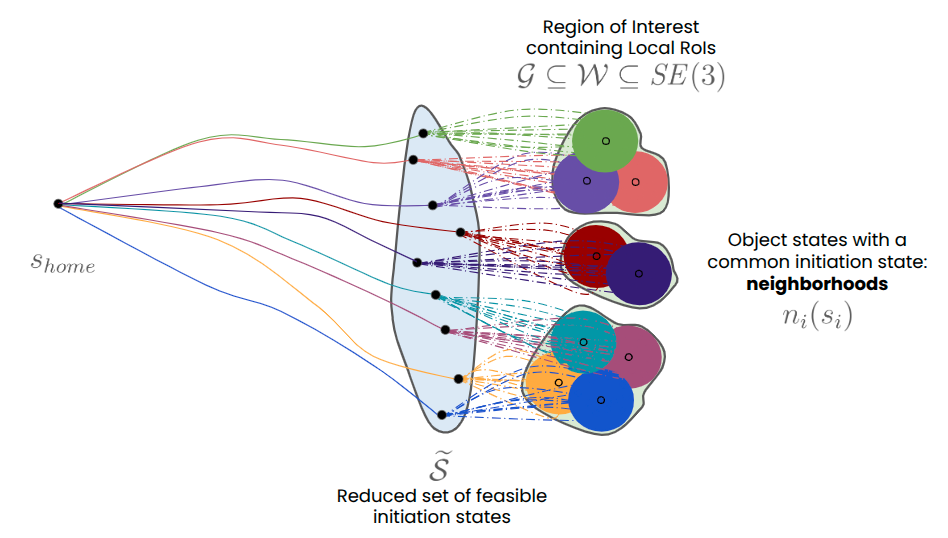}
    \caption{Overview of the preprocessing phase. Preprocessing computes a reduced set of initiation states $\widetilde{\mathcal{S}}$, each reached from $s_{home}$ by a precomputed path, and each covering a neighborhood $n_i(s_i)$ of object states within the region of interest $\mathcal{G}$.}
    \label{fig:neighbourhoods}
\end{figure}

We represent each neighborhood by an \emph{attractor tuple} $(w_{attr}, s_{i,attr}, r, \tau)$: an attractor object state $w_{attr}$, an attractor initiation state $s_{i,attr}$, a radius $r$, and a collision-free path $\tau$ from $s_{home}$ to $s_{i,attr}$ (Fig.~\ref{fig:neighbourhoods}). The tuple certifies that every object state $w$ with $d(w, w_{attr}) \le r$ is $\lambda$-covered by $s_{i,attr}$.

Algorithm~\ref{alg:preprocess} builds this library. Since $\mathcal{W} \subseteq SE(3)$ is continuous, polling every object state
in $\mathcal{G}$ is intractable; we therefore grow neighborhoods over a
discretization $\bar{\mathcal{G}} \subset \mathcal{G}$, whose resolution
is chosen with respect to the pose-error tolerance of $\sigma$ (Sections~\ref{sec:stat_guarantee}
and~\ref{sec:experiments}).

\textsc{Preprocess} treats each local RoI $\bar{\mathcal{G}}_i$ independently. While states remain that are neither covered nor known infeasible, it draws one of them as a candidate attractor $w_{attr}$ (Line~\ref{line:attr_obj}) and queries \textsc{GetInitStates} for $\mathcal{S}(w_{attr})$ (Line~\ref{line:init_state}). \textsc{ConstructNeighborhood} then attempts to grow a certified neighborhood around $w_{attr}$ from one of these initiation states (Line~\ref{line:construct}). by evaluating each candidate $s_i \in \mathcal{S}(w_{attr})$ in turn. It first asks $\mathcal{P}$ for a collision-free path $\tau_i$ from $s_{home}$ to $s_i$ and discards candidates the robot cannot reach (Lines~\ref{line:plan}--\ref{line:unreach}). It then tests whether $s_i$ $\lambda$-covers $w_{attr}$ itself, discarding those that fail this seed test (Lines~\ref{line:certify}--\ref{line:seedfail}). In both these cases, the discarded states are marked infeasible and reported as such at query time. Each surviving candidate is grown by \textsc{CertifiedExpand} (Line~\ref{line:expand}), which repeatedly tests object states on the frontier of the current neighborhood and halts at the first state that fails, returning the neighborhood $n_i$ and its radius $r_i$. Of the candidates that survive, the one with the largest radius is returned (Line~\ref{line:argmax}), since a wider neighborhood covers more object states per stored path. This returned neighborhood is marked covered and its attractor tuple is appended to $\mathcal{L}$ (Line~\ref{line:store}). Frontier states that failed are left uncovered and are drawn as attractors of their own neighborhoods in later iterations.


Every admission decision above is a \emph{certification test}. Because one rollout is not a reliable witness when $\pi_\sigma$ is stochastic, \textsc{CertifyRollouts} (Line~\ref{line:certify}) executes a batch of $m$ independent rollouts---fresh seeds, independently re-sampled noise---and returns a lower confidence bound $L$ on $p_\sigma$ at a per-test significance level $\delta$; the state under test is admitted iff $L \ge \lambda$. For deterministic behaviors, one successful rollout already identifies $p_\sigma \in \{0,1\} $, so $m = 1$ is sufficient for certification.

Each iteration of Algorithm~\ref{alg:preprocess} either admits
$w_{attr}$ to a stored neighborhood or marks it infeasible, so the set of unprocessed states shrinks by at least one per iteration; hence the algorithm terminates, and at most
$|\bar{\mathcal{G}}_i|$ attractors are sampled per local RoI. In the
absence of spatial coverage, preprocessing degrades to the naive
approach: every state in $\bar{\mathcal{G}}$ is sampled as an attractor
and tested against up to $\max_{w} |\mathcal{S}(w)|$ candidate
initiation states, and because each candidate attempts a neighborhood
expansion, every state in a local RoI may additionally be tested from
every other state in that RoI. Therefore, we can bound the certification tests in the preprocessing phase by:
\begin{equation}
  N_{\max} \;=\; |\bar{\mathcal{G}}| \cdot \max_{w} |\mathcal{S}(w)| \cdot \max_{i} |\bar{\mathcal{G}}_i|
  \label{eq:nmax}
\end{equation}
In practice, neighborhoods cover many states at once and fewer tests are executed; $N_{\max}$ enters only as the a~priori bound needed to set $\delta$ (Section~\ref{sec:stat_guarantee}).

\begin{algorithm}[t]
  \caption{Preprocess with Behaviors}
  \label{alg:preprocess}
  \footnotesize
  \SetAlgoLined\DontPrintSemicolon
  \SetKwProg{Fn}{Procedure}{:}{}
  \SetKwFunction{Preprocess}{\scriptsize Preprocess}
  \SetKwFunction{Sample}{\scriptsize Sample}
  \SetKwFunction{GetInitStates}{\scriptsize GetInitStates}
  \SetKwFunction{ConstructNeighborhood}{\scriptsize ConstructNeighborhood}
  \SetKwFunction{PlanPath}{\scriptsize PlanPath}
  \SetKwFunction{CertifyRollouts}{\scriptsize CertifyRollouts}
  \SetKwFunction{CertifiedExpand}{\scriptsize CertifiedExpand}
  \SetKw{KwContinue}{continue}
  \KwIn{$s_{home}$: robot home state\newline
        $\mathcal{P}$: collision-free motion planner\newline
        $\bar{\mathcal{G}} = \bigcup_i \bar{\mathcal{G}}_i$: discretized RoI\newline
        $\sigma$: manipulation behavior\newline
        $\lambda$: success-rate threshold\newline
        $m, \delta$: certification batch size and per-test significance level, $\delta = \alpha / N_{\max}$ (Section~\ref{sec:stat_guarantee})}
  \KwOut{Library $\mathcal{L}$ of attractor tuples $(w_{attr}, s_{i,attr}, r, \tau)$}
  \Fn{\Preprocess($s_{home}, \mathcal{P}, \bar{\mathcal{G}}, \sigma, \lambda, m, \delta$)}{
    $\mathcal{L} \gets \emptyset$\;
    \ForEach{$\bar{\mathcal{G}}_i \in \bar{\mathcal{G}}$}{
      $\bar{\mathcal{G}}_i^{cov}, \bar{\mathcal{G}}_i^{inf} \gets \emptyset$\;
      \While{$\bar{\mathcal{G}}_i \setminus (\bar{\mathcal{G}}_i^{cov} \cup \bar{\mathcal{G}}_i^{inf}) \neq \emptyset$}{
        $w_{attr} \gets$ \Sample($\bar{\mathcal{G}}_i \setminus (\bar{\mathcal{G}}_i^{cov} \cup \bar{\mathcal{G}}_i^{inf})$)\label{line:attr_obj}\;
        $\mathcal{S}(w_{attr}) \gets$ \GetInitStates($w_{attr}$)\label{line:init_state}\;
        $(n, r, s_{i,attr}, \tau) \gets$ \ConstructNeighborhood($s_{home}, w_{attr}, \mathcal{S}(w_{attr}), \mathcal{P}, \sigma, \lambda, m, \delta$)\label{line:construct}\;
        \lIf{$s_{i,attr} = \bot$}{\label{line:infeasible}$\bar{\mathcal{G}}_i^{inf} \gets \bar{\mathcal{G}}_i^{inf} \cup \{w_{attr}\}$; \KwContinue \Comment*[h]{not statistically $\lambda$-feasible (Definition~\ref{def:stat_coverage})}}
        $\bar{\mathcal{G}}_i^{cov} \gets \bar{\mathcal{G}}_i^{cov} \cup n$;\quad $\mathcal{L} \gets \mathcal{L} \cup \{(w_{attr}, s_{i,attr}, r, \tau)\}$\label{line:store}\;
      }
    }
    \Return{$\mathcal{L}$}\;
  }
  \vspace{1mm}
  \Fn{\ConstructNeighborhood($s_{home}, w_{attr}, \mathcal{S}(w_{attr}), \mathcal{P}, \sigma, \lambda, m, \delta$)}{
    $C \gets \emptyset$ \Comment*[h]{certified candidates}\;
    \ForEach{$s_i \in \mathcal{S}(w_{attr})$}{
      $\tau_i \gets \mathcal{P}.\PlanPath(s_{home}, s_i)$\label{line:plan}\;
      \lIf{$\tau_i = \bot$}{\label{line:unreach}\KwContinue \Comment*[h]{unreachable}}
      $L \gets$ \CertifyRollouts$(s_i, w_{attr}, \sigma, m, \delta)$\label{line:certify} \Comment*[h]{$m$ rollouts; lower bound on $p_\sigma(s_i, w_{attr})$}\;
      \lIf{$L < \lambda$}{\label{line:seedfail}\KwContinue \Comment*[h]{seed not certified}}
      $(n_i, r_i) \gets$ \CertifiedExpand$(w_{attr}, s_i, \sigma, \lambda, m, \delta)$\label{line:expand} \Comment*[h]{grow while frontier certifies}\;
      $C \gets C \cup \{(n_i, r_i, s_i, \tau_i)\}$\;
    }
    \lIf{$C = \emptyset$}{\Return{$(\emptyset, 0, \bot, \bot)$}}
    \Return{$\argmax_{(n, r, s, \tau)\, \in\, C} r$}\label{line:argmax} \Comment*[h]{widest certified neighborhood}
  }
\end{algorithm}

\begin{algorithm}[t]
  \caption{Query}
  \label{alg:query}
  \footnotesize
  \SetAlgoLined\DontPrintSemicolon
  \SetKwProg{Fn}{Procedure}{:}{}
  \SetKwFunction{Query}{\scriptsize Query}
  \SetKwFunction{Lookup}{\scriptsize Lookup}
  \SetKwFunction{Execute}{\scriptsize Execute}
  \SetKwFunction{Rollout}{\scriptsize Rollout}
  \KwIn{$\mathcal{R}$: robot, at $s_{home}$\newline
        $w_g \in \mathcal{G}$: queried object state\newline
        $\mathcal{L}$: preprocessed library\newline
        $\sigma$: manipulation behavior}
  \Fn{\Query($\mathcal{R}, w_g, \mathcal{L}, \sigma$)}{
    $(w_{attr}, s_{i,attr}, r, \tau) \gets$ \Lookup($\mathcal{L}, w_g$)\label{line:lookup} \Comment*[h]{tuple with $d(w_g, w_{attr}) \le r$, or $\bot$ if none}\;
    \lIf{$\tau = \bot$}{\Return{\textit{infeasible}}}
    $\mathcal{R}$.\Execute($\tau$)\label{line:execute} \Comment*[h]{collision-free motion $s_{home} \to s_{i,attr}$}\;
    $\mathcal{R}$.\Rollout($\sigma, s_{i,attr}, w_g$)\label{line:rollout} \Comment*[h]{execute $\pi_\sigma$ from $s_{i,attr}$; succeeds w.p.\ $\ge \lambda$}\;
  }
\end{algorithm}

\subsection{Online Phase}
\label{alg_app:online}
At the successful completion of the preprocessing phase, we obtain a library $\mathcal{L}$ of stored attractor-tuples.
This library enables fast online queries when a goal object state $w_g \in \mathcal{G}$ becomes available.
Given a query $w_g \in \mathcal{G}$, the online phase proceeds in three steps, highlighted in Algorithm~\ref{alg:query}. The first is \textit{region identification} (Line~\ref{line:lookup}): we retrieve the attractor tuple whose neighborhood contains $w_g$, i.e.\ the one satisfying $d(w_g, w_{attr}) \leq r$ under the chosen metric $d(\cdot,\cdot)$. If no such tuple exists, $w_g$ was marked infeasible during preprocessing and is reported as such. The second step executes the stored collision-free path $\tau$, which carries the robot from $s_{home}$ to the attractor initiation state $s_{i,attr}$ (Line~\ref{line:execute}). The third invokes the behavior rollout from $s_{i,attr}$, executing $\pi_\sigma$ to complete the task with certified success probability at least $\lambda$ (Line~\ref{line:rollout}).

Hence, the online phase reduces to simple lookup operations and direct plan execution, ensuring consistent performance within the desired time bound $T_{bound}$.

%% file: sections/stat_guarantee.tex
\subsection{Theoretical Guarantees}
\label{sec:stat_guarantee}

\subsubsection{Constant Time Guarantee}
An online query is a lookup over the preprocessed library $\mathcal{L}$,
identifying which neighborhood contains the target object state. Its cost is
therefore independent of the query, bounded by a scan of $\mathcal{L}$, whose
size is fixed at preprocessing time and does not grow with the query; the
work per query is thus $O(1)$ in the query itself, and $T_{\mathrm{bound}}$
depends only on $|\mathcal{L}|$, which is fixed at preprocessing time.  Hence, by
Definition~\ref{def:constant_time}, B-CTMP is a constant-time planner for all certified
$\lambda$-feasible goal states in $\bar{\mathcal{G}}$.

\subsubsection{Statistical Certification}

Given a user-specified success threshold $\lambda$, our aim is to guarantee
that every object state in every neighborhood stored by B-CTMP is
$\lambda$-covered by that neighborhood's attractor initiation state---that
is, that the behavior, rolled out from the stored initiation state, succeeds
with probability at least $\lambda$. Because the true success probability
function $p_\sigma(s_i,w)$ is unknown, this coverage cannot be verified
directly and must instead be established statistically, by means of a
\emph{certification test}, which decides whether an object state $w$ belongs to
the neighborhood $n_i(s_i)$. In this
test, a batch of $m$ independent rollouts of $\pi_\sigma$ from $s_i$ is
executed, each with a freshly drawn random seed and independently
re-sampled noise. The rollout outcomes are therefore i.i.d.\ Bernoulli
trials with unknown true success rate $p := p_\sigma(s_i, w)$.

We can now apply standard statistical machinery to turn this batch of
outcomes into a decision. Let $\delta \in (0,1)$ be
the \textbf{per-test significance level} at which a single test is required
to certify. From these outcomes we compute the Clopper--Pearson lower
confidence bound $L$ on $p$ at level $\delta$~\cite{clopper1934use}.
The number of rollouts $m$ is a design parameter that governs the tightness
of this bound: for a fixed $\delta$, the gap between $L$ and the empirical
success rate shrinks as $m$ grows, so a larger $m$ allows a pair to be
certified whose true success probability lies closer to $\lambda$, at the
cost of proportionally more preprocessing time. Coverage is determined by
comparing $L$ against $\lambda$, which leads to the following
definition:

\begin{mydef}[Statistical $\lambda$-Coverage]
\label{def:stat_coverage}
An initiation state $s_i$ \emph{statistically $\lambda$-covers} an object
state $w$ if $L \ge \lambda$ for a certification test performed on the pair
$(s_i, w)$. Since $L$ satisfies $\Pr[\, p < L \,] \le \delta$ for every
$p \in [0,1]$, statistical $\lambda$-coverage certifies, with confidence
$1-\delta$, that $p := p_\sigma(s_i, w) \ge \lambda$. Such object states are termed statistically $\lambda$-feasible.
\end{mydef}

In order to extend this guarantee to the entire RoI, we must choose
$\delta$ to support a user-specified \textbf{significance level}
$\alpha \in (0,1)$. Recall from Section~\ref{alg_app:preprocess} that preprocessing
executes at most $N_{\max}$ certification tests. We therefore set
$\delta := \alpha / N_{\max}$ for every test, which allows us to state the
following theorem:

\begin{theorem}[RoI-Wide Soundness]
\label{thm:soundness}
With probability at least $1-\alpha$, every object state admitted into any
neighborhood stored during preprocessing is $\lambda$-covered (Definition~\ref{def:lam_cover}) by its attractor initiation state.
\end{theorem}

\begin{proof}
The proof is by construction. Let $E_j$ denote the event that the $j$-th
certification test admits an object state that is not $\lambda$-covered.
By Definition~\ref{def:stat_coverage}, $\Pr[E_j] \le \delta$ for each $j$.
The event that \emph{any} test fails is the union of these $N_{\max}$
events, so by the union bound (equivalently, the Bonferroni
correction~\cite{bonferroni}),
\begin{equation}
    \Pr\Big[\bigcup_{j=1}^{N_{\max}} E_j\Big]
    \;\le\; \sum_{j=1}^{N_{\max}} \Pr[E_j]
    \;\le\; N_{\max}\,\delta
    \;=\; N_{\max} \cdot \frac{\alpha}{N_{\max}}
\end{equation}
Hence, with probability at least $1-\alpha$, no test fails and every
admitted object state is $\lambda$-covered by its attractor initiation
state.
\end{proof}

Thus, B-CTMP statistically certifies the goal RoI. 


\subsubsection{Completeness} To characterize B-CTMP’s solution guarantees, we introduce a notion of completeness tailored to behavior-based
manipulation planning.


\begin{mydef}[$(\alpha)$-Completeness]
\label{def:alpha_complete}
An algorithm is \emph{$(\alpha)$-complete} over $\bar{\mathcal{G}}$ if for every statistically
$\lambda$-feasible object state $w \in \bar{\mathcal{G}}$
(Definition~\ref{def:lambda_feasible}), the algorithm returns a solution
$\xi = (\tau, \pi_\sigma)$ whose collision-free path $\tau$ terminates
at an initiation state that $\lambda$-covers $w$
(Definition~\ref{def:lam_cover}); otherwise, it reports that no plan
exists for the given behavior, goal, and success threshold. Since statistical $\lambda$-feasibility holds with probability $1-\alpha$, we characterize completeness with respect to that value as well. 
\end{mydef}

Both $\lambda$ and $\alpha$ are specified by the user according to the
requirements of the task at hand.

\begin{remark}
$(\alpha)$-Completeness is relative to \textsc{GetInitStates}, by the construction of the behavior \ref{sec:preliminary}. An object state servable only from an initiation state outside $\mathcal{S}(w)$ is therefore reported infeasible, so the guarantee inherits the coverage of \textsc{GetInitStates}.
\end{remark}

\begin{theorem}[B-CTMP $(\alpha)$-Completeness]
\label{thm:a-complete}
B-CTMP with certified preprocessing (Algorithm~\ref{alg:preprocess}) is $(\alpha)$-complete within the preprocessed region-of-interest $\bar{\mathcal{G}}$.
\end{theorem}

\begin{proof}[Proof sketch]
Algorithm~\ref{alg:preprocess} samples attractors from each local RoI
until $\bar{\mathcal{G}}_i \setminus (\bar{\mathcal{G}}_i^{cov} \cup
\bar{\mathcal{G}}_i^{inf}) = \emptyset$, so at termination every
$w \in \bar{\mathcal{G}}$ lies in a stored neighborhood or in
$\bar{\mathcal{G}}^{inf}$. We argue on the outcome of the online lookup.

If the lookup succeeds, it returns a tuple
$(w_{attr}, s_{attr}, r, \tau) \in \mathcal{L}$ with
$d(w_g, w_{attr}) \le r$, where $s_{attr}$ was certified on $w_g$ during
expansion. By Theorem~\ref{thm:soundness}, $s_{attr}$
$\lambda$-covers $w_g$. If the lookup fails, then $w_g$ lies in no stored neighborhood, hence
$w_g \in \bar{\mathcal{G}}^{inf}$ by the termination condition. It
remains to show that this is the correct verdict. A state enters
$\bar{\mathcal{G}}^{inf}$ only after being selected as an attractor
itself, at which point its initiation set $\mathcal{S}(w_g)$ is queried
via \textsc{GetInitStates} and every $s_i \in \mathcal{S}(w_g)$ is
tested for a collision-free path from $s_{home}$ and for
$\lambda$-certification. This test is decisive: if every $s_i$ fails,
then no initiation state of $w_g$ is both reachable and $\lambda$-covering,
which is precisely the negation of $\lambda$-feasibility
(Definition~\ref{def:lambda_feasible}), and excluding $w_g$ from
$\mathcal{L}$ is the required behavior. Conversely, a state at which
\textsc{CertifiedExpand} halts while growing some \emph{other}
attractor's neighborhood is never marked infeasible on that basis: it
remains uncovered and is later drawn as an attractor in its own right.
Hence no state is rejected for merely falling outside a neighbor's
certified region. Thus B-CTMP is $(\alpha)$-complete within
$\bar{\mathcal{G}}$.
\end{proof}

%% file: sections/experiments.tex
\section{Experiments}
\label{sec:experiments}
We evaluate B-CTMP on three manipulation tasks---shelf picking, plug insertion, and wheel replacement---in simulation, and demonstrate on real UR10e and Kinova robots. The tasks are chosen to span both deterministic and stochastic regimes of the behavior model. Through these experiments we validate the central claims of B-CTMP proved conceptually in section \ref{sec:stat_guarantee}.
\subsection{Manipulation Tasks}

\subsubsection{Jacobian-Based Behaviors}
The first two tasks share a common behavior structure: the target object pose $w_g \in \mathrm{SE}(3)$ is estimated from perception and specified in the world frame, and a closed-loop policy is rolled out from an initiation state to a goal pose computed from $w_g$ and the known object model, at which point a task-specific terminal action is executed. For both tasks, the distance metric $d$ is the Euclidean distance between the query object pose and the attractor object pose in $(x, y, \theta)$,
since the static geometry restricts the object's movement in the
remaining directions. The two tasks differ in their scenario and in how \textsc{GetInitStates} proposes initiation states:

\noindent\textbf{Task 1: Shelf Grasping.} Modelling a scenario commonly encountered in industrial warehouse automation, we consider a manipulator that must retrieve a target object from structured storage. The grasp behavior $\sigma_{grasp}$ is defined by a pre-grasp pose (the initiation state) and a grasp pose, where a closure sequence is activated. \textsc{GetInitStates} computes candidate grasp poses with valid IK solutions from the known object model, scores them by an antipodal metric derived from local surface normals, and retains the top $K$. Pre-grasp poses are then obtained by applying a fixed retraction transformation along each grasp's approach.

\noindent\textbf{Task 2: Charger Insertion.} We consider a task commonly required in automated charging systems and electronic assembly, where a connector must be inserted into a port at $w_g$ under strict geometric constraints. Since the port geometry admits a single insertion pose, \textsc{GetInitStates} obtains the pre-insert initiation state by applying a fixed retraction to that pose.

\noindent\textbf{Behavior Policy.} In both tasks, the policy uses Jacobian control to move the end-effector through differential kinematics from the initiation state to the goal pose. To ensure robustness under perception uncertainty, we impose a manipulability constraint during policy execution: we define the minimum manipulability radius as $r_{\min}(q) = \min_i \sqrt{\lambda_i\!\big(J(q) J(q)^{\mathsf T}\big)}$, where the manipulability ellipsoid radii are derived from the Jacobian eigenvalues, and mark a rollout infeasible if $r_{\min}(q) < \epsilon$, where $\epsilon$ is the perception noise bound\footnote{The perception noise bound $\epsilon$ is determined from the camera calibration process, which provides an upper bound on the pose estimation error.}. This ensures that stored initiation states retain adequate dexterity for corrective motions across their neighborhoods. Since the closed-loop trajectory is fully determined by the initiation state and object pose, both behaviors are deterministic: $p_\sigma(s_i, w) \in \{0,1\}$, and certification uses a single rollout.

\subsubsection{Learned Behavior} For the third task, the behavior is executed by a learned policy.

\noindent\textbf{Task 3: Wheel Replacement.} We consider mounting a wheel onto a hub, a representative assembly operation in automotive maintenance and manufacturing. It requires seating the wheel's bolt holes onto the hub studs at $w_g \in \mathrm{SE}(3)$, estimated from perception as before. The distance metric $d$ is the Euclidean distance between the query and attractor hub poses in $(x, y)$, since a mounted hub's orientation is fixed by the vehicle geometry and only its position varies across instances. Since the hub geometry fixes a unique mounting pose, \textsc{GetInitStates} applies a fixed retraction along the stud axis and solves randomized IK to yield $K$ distinct initiation states.

\noindent\textbf{Behavior Policy.} The behavior policy $\pi_{\sigma}$ is trained with PPO \cite{schulman2017ppo} in
ManiSkill \cite{taomaniskill3}, from scratch on the randomized-configuration wheel-mount task, for approximately $19$ hours. We utilize a dense wheel–stud geometric reward: lateral alignment ($\leq 3.0$), hole–stud proximity ($\leq 2.5$), insertion depth ($\leq 2.5$), and uprightness ($\leq1.0$), totaling $9.0$. Proximity is discounted by tilt and misalignment with \(8^\circ\) and \(5\,\mathrm{mm}\) scales. Successful insertion receives up to $10.0$. From the initiation state, the policy closes the loop on the end-effector pose relative to the goal frame to align and seat the wheel; a rollout is successful if every bolt hole is within $\epsilon$ of its nearest stud along the stud axis.

\subsection{Baseline Methods}
\label{sec:baselines}
Our baselines are chosen to isolate the three claims of Section~\ref{sec:experiments}. The first family---fully online planning---is the standard practice for chaining collision-free motion to a behavior, and tests both planning speed and whether initiation states found on the fly support successful execution. The second family---preprocessing-based planning---also achieves fast online queries through offline computation, but without behavior awareness. Comparing against it tests whether B-CTMP's reliability comes from speedup alone or from the statistical machinery used in preprocessing. For learned behaviors, a third baseline---an end-to-end policy trained to complete the task directly from $s_{\text{home}}$---tests whether the two-phase structure that B-CTMP certifies is necessary at all, or whether a single policy trained over the full workspace can subsume both phases. 

All methods share the same \textsc{GetInitStates}, behavior policy $\pi_\sigma$, and motion planner $\mathcal{P}$; since B-CTMP treats each as a black box, improvements to any of them (e.g., learned initiation-state proposers) would benefit every method equally, so we hold them fixed and vary only how motion and behavior are coupled.

\noindent\textbf{Online Planning Baseline:} For each query $w_g$, this baseline polls \textsc{GetInitStates} to obtain $\mathcal{S}(w_g)$, plans a collision-free motion to each initiation state in turn, and executes $\pi_\sigma$ from the terminal state of the first plan found. A ``re-planning'' variant instead continues through $\mathcal{S}(w_g)$, repeating motion planning and rollout until the behavior succeeds or the time budget is exhausted. Both are implemented with BiTRRT \cite{devaurs2013bitrtt} from OMPL.

\noindent\textbf{Preprocessing-based Baselines:} We implement two variants that leverage offline computation to accelerate online planning. The first utilizes an offline PRM graph that seeds the planner with a precomputed roadmap. The second employs prior CTMP methods \cite{mishani2024constant} to store a library of paths to a manually defined initiation region $\mathcal{S}_{\text{init}}$, estimated to at least partially overlap with $I_\sigma(w_g)$ for different target poses $w_g$. Both follow the same online protocol: they poll \textsc{GetInitStates} to obtain $\mathcal{S}(w_g)$, then connect each initiation state to their precomputed structure through method-specific means, with details deferred to the respective methods. Finally, both execute $\pi_\sigma$ from the resulting path's terminal state.

\noindent\textbf{Learned Global Policy:} For the learned behavior, we additionally train an end-to-end policy that maps directly from $s_{\text{home}}$ and $w_g$ to task completion. 

\noindent\textbf{Success Criterion:} For all methods, we execute $m$ batched rollouts of $\pi_\sigma$ from the terminal state of each returned plan and count the query as a success only if the resulting lower confidence bound satisfies $L \geq \lambda$, in order to validate statistical performance on the query.


%
\newlength{\rowgutter}
\newlength{\imgw}
\newlength{\tabw}
\setlength{\rowgutter}{6pt}

%
\newlength{\tabslack}
\setlength{\tabslack}{6pt}
\setlength{\rowgutter}{8pt}
\newsavebox{\tabbox}
\newcommand{\fittab}[1]{%
  \sbox{\tabbox}{#1}%
  \ifdim\wd\tabbox>\tabw
    \resizebox{\tabw}{!}{\usebox{\tabbox}}%
  \else
    \usebox{\tabbox}%
  \fi}
\newcommand{\setrowwidths}{%
  \setlength{\imgw}{0.50\linewidth}%
  \setlength{\tabw}{\linewidth}%
  \addtolength{\tabw}{-\imgw}%
  \addtolength{\tabw}{-\rowgutter}%
  \addtolength{\tabw}{-\tabslack}%
}
\newcommand{\rowtitle}[1]{%
  \par\vspace{7pt}%
  {\centering\textsc{#1}\par}%
  \vspace{3pt}}

\begin{figure*}[t]
  \centering
  \setrowwidths
  \setlength{\tabcolsep}{3pt}
  \renewcommand{\arraystretch}{1.1}

  \rowtitle{Shelf Grasping}
  \noindent
  \begin{minipage}[c]{\imgw}%
    \centering
    \includegraphics[width=\linewidth]{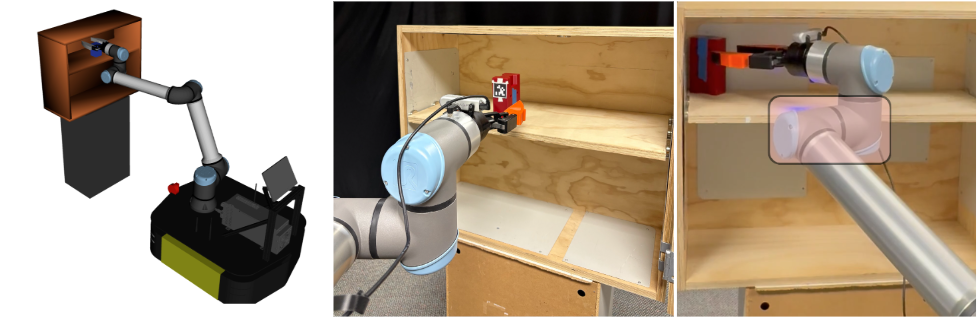}%
  \end{minipage}%
  \hspace{\rowgutter}%
  \begin{minipage}[c]{\tabw}%
    \centering\footnotesize
    \fittab{\begin{tabular}{@{}lccc@{}}
      \toprule
      \textbf{Method} & \textbf{End-to-End} $\uparrow$ & \textbf{Path Only} & \textbf{Planning Time} $\downarrow$ \\
                      & [\%] & (rollout failed) [\%] & [ms] \\
      \midrule
      PRM                    & 72.7 & 27.3 & $2780 \pm 170$ \\
      BiTRRT                 & 67.7 & 27.3 & $2700 \pm 300$ \\
      BiTRRT (Replan)        & 73.0 & 27.1 & $3200 \pm 400$ \\
      Vanilla CTMP           & 59.6 & 23.2 & $328 \pm 30$   \\
      \textbf{B-CTMP (Ours)} & $\bm{100.0}$ & $0.0$ & $\bm{0.600 \pm 0.100}$ \\
      \bottomrule
    \end{tabular}}%
  \end{minipage}%

  \rowtitle{Charger Insertion}
  \noindent
  \begin{minipage}[c]{\imgw}%
    \centering
    \includegraphics[width=\linewidth]{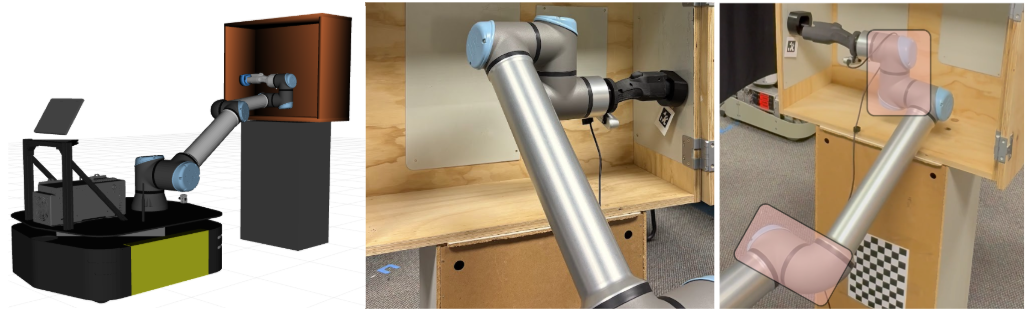}%
  \end{minipage}%
  \hspace{\rowgutter}%
  \begin{minipage}[c]{\tabw}%
    \centering\footnotesize
    \fittab{\begin{tabular}{@{}lccc@{}}
      \toprule
      \textbf{Method} & \textbf{End-to-End} $\uparrow$ & \textbf{Path Only} & \textbf{Planning Time} $\downarrow$ \\
                      & [\%] & (rollout failed) [\%] & [ms] \\
      \midrule
      PRM                    & 51.6 & 5.9  & $2520 \pm 310$ \\
      BiTRRT                 & 38.0 & 10.5 & $2500 \pm 320$ \\
      BiTRRT (Replan)        & 23.0 & 62.9 & $3024 \pm 500$ \\
      Vanilla CTMP           & 65.0 & 23.3 & $440 \pm 150$  \\
      \textbf{B-CTMP (Ours)} & $\bm{100.0}$ & $0.0$ & $\bm{0.900 \pm 0.300}$ \\
      \bottomrule
    \end{tabular}}%
  \end{minipage}%

  \rowtitle{Wheel Insert}
  \noindent
  \begin{minipage}[c]{\imgw}%
    \centering
    \includegraphics[width=\linewidth]{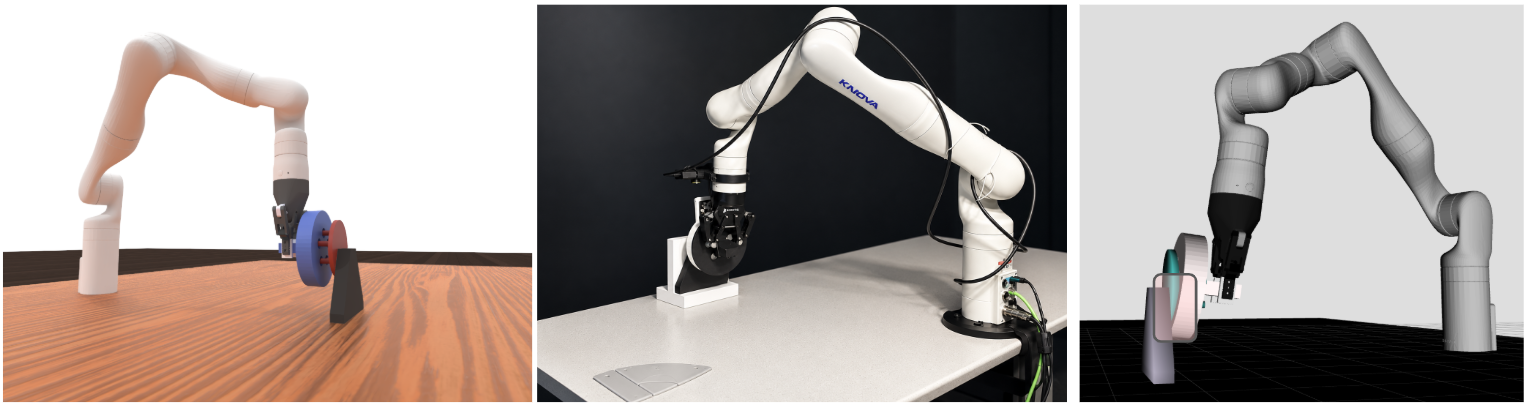}%
  \end{minipage}%
  \hspace{\rowgutter}%
  \begin{minipage}[c]{\tabw}%
    \centering\footnotesize
    \fittab{\begin{tabular}{@{}lccc@{}}
      \toprule
      \textbf{Method} & \textbf{End-to-End} $\uparrow$ & \textbf{Path Only} & \textbf{Planning Time} $\downarrow$ \\
                      & [\%] & (rollout failed) [\%] & [ms] \\
      \midrule
      PRM                    & 88.0 & 8.0  & $1180 \pm 10$ \\
      BiTRRT                 & 93.0 & $0.0$ & $1176 \pm 3$ \\
      BiTRRT (Replan)        & 93.0 & $0.0$ & $1176 \pm 10.5$ \\
      Vanilla CTMP           & 92.5 & 7.5 & $1003 \pm 100$ \\
      Learned Global Policy          & 58 & - & - \\
      \textbf{B-CTMP (Ours)} & $\bm{99.5}$ & $0.5$ & $\bm{0.041 \pm 0.004}$ \\
      \bottomrule
    \end{tabular}}%
  \end{minipage}%

  \caption{Experimental results across the three manipulation tasks. Each table reports (i) the end-to-end success rate over feasible task instances ($N{=}100$ grasping, $N{=}60$ insertion, and $N{=}200$ wheel replacement queries), scored under the $L \geq 0.9$ criterion of Section~\ref{sec:baselines}, and (ii) the online planning time. Left panels show representative infeasible queries. For grasping (top), the object is placed where every candidate grasp pose collides with static obstacles. For insertion (middle), the port pose induces a joint singularity, causing loss of manipulability and failure to reach the target. For wheel replacement (bottom), the hub pose lies in a location where the policy results in a collision. Red bounding boxes indicate the affected joints.}
  \label{fig:results_main}
\end{figure*}

\subsection{Experimental Setup}
All plans start from the robot home state $s_{home}$, with a 5\,s timeout for every planner. Goal poses are sampled i.i.d.\ from the same region of interest used during preprocessing to cover diverse regions of each task space, spanning varying spatial localities, manipulation complexities, and initiation states. The discretization resolution is $\Delta = 2\,\mathrm{cm}$ in all three tasks. We set the success threshold to $\lambda=0.9$ and the RoI-wide significance to $\alpha=0.05$. For the learned behavior the batch size is bounded by available compute, and we set $m=256$, the largest batch that fits in a single parallel rollout. Parallel rollouts and policy training run on an NVIDIA RTX 5090 GPU, and all planning-time results are measured on an Intel Core Ultra 9 185H CPU. We demonstrate batched rollout based quantitative comparisons in simulation. We additionally evaluate B-CTMP on real robots—a UR10e for grasping and plug insertion, and a Kinova for wheel replacement—running the behavior once per trial to mimic realistic deployment and evaluate execution reliability.


\subsection{Results and Analysis}
\subsubsection{Planning Performance}
Fig.~\ref{fig:results_main} presents the success rate and planning time for B-CTMP compared to the baseline methods in simulation. All success rates for both the baselines and B-CTMP are reported on $\lambda$-feasible instances only, since infeasible object states always fail on the baselines and are reported as such by B-CTMP. This rejection is itself a safety property: baselines attempt execution from such states, which on hardware drives the arm into singularities or collisions (Fig.~\ref{fig:results_main}), whereas B-CTMP declines them in constant time. In both experiments, our planner maintained a 100\% end-to-end success rate, consistent with the certified lower bound on the composite plan's success probability, whereas the baselines failed either at finding valid collision-free paths to initiation states or during behavior execution from the path terminal states. Our approach also demonstrates sub-millisecond online query performance through fast lookup operations, compared to expensive online computation required by baseline methods. 

\subsubsection{Baseline Failure Mode Analysis}
The baselines fail in three distinct ways. Most failures stem from treating behavior validation as an afterthought: because motion planning is decoupled from execution, the planner reaches initiation states that are kinematically valid but behaviorally invalid. In the insertion domain, shelf corners and side walls proved especially difficult, yielding initiation states with poor manipulability or unfavorable approach angles. Second, none of the baselines reason about stochastic execution, so a state that succeeds once may fail on the next attempt and no guarantee can be offered. Third, the end-to-end policy struggles over the long horizon from $s_{home}$, whereas the local policy performs far better: it need only succeed near its initiation state, with global reachability delegated to the certified motion plan.

\begin{table}[t]
  \centering
  \footnotesize
  \setlength{\tabcolsep}{4.5pt}
  \renewcommand{\arraystretch}{0.9}
  \setlength{\aboverulesep}{0.2ex}
  \setlength{\belowrulesep}{0.3ex}
  \caption{Real-robot experiments. We report preprocessing statistics, showing the growth of preprocessing time with goal-region volume and the memory compression relative to a naive baseline storing every precomputed path. For each task we also report B-CTMP's coverage, as the fraction of queries declined as infeasible, and the execution success rate on feasible queries (50, 50, and 20 trials, respectively).}
  \label{tab:results}
  \begin{tabular}{@{}l
      S[table-format=5.0]
      S[table-format=2.2, round-mode=places, round-precision=2]
      S[table-format=2.2, round-mode=places, round-precision=2]@{}}
    \toprule
    \multirow{2}{*}{\textbf{Task}} & {\textbf{Goal Region}} & {\textbf{Preprocessing}} & {\textbf{Memory}} \\
                                   & {\textbf{Vol. [cm$^3$]}} & {\textbf{Time [h]}} & {\textbf{Compression [\%]}} \\
    \midrule
    \multirow{5}{*}{Shelf Grasping}    & 1260  & 0.28  & 99.6  \\
                                       & 2500  & 5.90  & 94.0  \\
                                       & 5600  & 14.5  & 92.7  \\
    \cmidrule(l){2-4}
                                       & {\textbf{Succ. [\%]}} & {\textbf{Infeas. [\%]}} & {\textbf{Plan. [ms]}} \\
                                       & {100} & {18} & {1.8\, $\pm$ \, 0.007} \\
    \midrule
    \multirow{5}{*}{Charger Insertion} & 21000 & 0.09  & 56.0  \\
                                       & 45000 & 0.17  & 61.8  \\
                                       & 96000 & 0.45  & 67.1  \\
    \cmidrule(l){2-4}
                                       & {\textbf{Succ. [\%]}} & {\textbf{Infeas. [\%]}} & {\textbf{Plan. [ms]}} \\
                                       & {100} & {20} & {1.3\, $\pm$\, 0.05} \\
    \midrule
    \multirow{5}{*}{Wheel Replacement} & 6160  & 4.46  & 98.07 \\
                                       & 12320 & 9.36  & 97.55 \\
                                       & 24000 & 22.4  & 97.44 \\
    \cmidrule(l){2-4}
                                       & {\textbf{Succ. [\%]}} & {\textbf{Infeas. [\%]}} & {\textbf{Plan. [ms]}} \\
                                       & {100} & {15} & {0.15\,$\pm$\,0.11} \\
    \bottomrule
  \end{tabular}
\end{table}

\subsubsection{Preprocessing Analysis}

Table~\ref{tab:results} reports preprocessing time and memory compression across goal region volumes for each task. Memory compression is measured relative to a naive baseline that explicitly computes and stores an individual path to a feasible initiation state for every discretized object pose in the goal region. B-CTMP reduces memory by over 92\% for shelf grasping, over 97\% for wheel replacement, and 56--67\% for charger insertion. This reduction directly benefits online planning, since plan retrieval time scales with cache size and excessive memory overhead can introduce latency during real-time execution.

Preprocessing cost depends on more than goal region volume. For example, shelf grasping takes considerably longer than charger insertion, because insertion admits only one initiation state per object pose and therefore requires fewer rollouts and planning attempts. Cost also grows with the certification parameters, since a higher $\lambda$ or a larger batch size $m$ means more rollouts per test. 


\subsubsection{Real-Robot Performance}
Table~\ref{tab:results} also reports real-robot results, which we analyze along two axes: \emph{coverage}, the fraction of queries for which B-CTMP returns a plan, and \emph{execution reliability}, the fraction of returned plans whose behavior succeeds. Queries B-CTMP declines to cover are listed in the infeasible column; representative cases are shown in Fig.~\ref{fig:results_main}. On covered queries, execution succeeded in every trial under real perception noise, since each stored motion terminates at an initiation state certified against a high success threshold during preprocessing. Real robot demonstrations and representative infeasible query executions can be seen in the supplementary video.


%% file: sections/conclusion_future_work.tex


\vspace{-1mm}
\section{Conclusion and Limitations}
We introduced B-CTMP, a constant-time planner that validates manipulation behaviors during preprocessing rather than treating motion planning and behavior execution as decoupled, sequential steps. Its time bound therefore covers completing the task, not merely reaching a robot configuration. Since such behaviors are stochastic, and one rollout is no evidence of success, validation takes the form of statistical certification: every cached plan is certified to succeed with probability at least $\lambda$, simultaneously across the cache at confidence $1-\alpha$; a single rollout recovers prior CTMP's reachability check. A second key is defining the region-of-interest in object space rather than configuration or task space, which automatically discovers relevant initiation states and eliminates the need for manual specification. Experiments show consistent success rates and sub-millisecond queries where baselines fail.
These guarantees rest on a few assumptions that define B-CTMP's scope. First, the environment is semi-static, with all geometry other than the object of interest fixed at preprocessing time (as in the warehouse and assembly settings we target). Second, each plan chains a collision-free motion to a single behavior- extension to behavior sequences is a promising direction for future work. Third, certification requires a simulator that faithfully reproduces execution, so model uncertainty and the sim-to-real gap may weaken offline guarantees. Finally, certification is paid for offline, with preprocessing ranging from minutes to hours, making B-CTMP best suited to repetitive, high-throughput tasks. Within this scope, B-CTMP provides a robust building block for predictable, behavior-aware robotic automation in semi-structured environments.